\pdfoutput=1

\documentclass[11pt]{article}

\usepackage[hyperref]{emnlp2021}
\usepackage{booktabs}
\usepackage{subcaption}

\usepackage{times}
\usepackage{latexsym}
\usepackage{amsmath}
\usepackage{amsfonts,amssymb}
\usepackage{bm}
\usepackage{tabularx}
\usepackage{lipsum}
\usepackage{amsthm}

\usepackage{multirow, multicol}
\usepackage{arydshln}
\usepackage[ruled,vlined]{algorithm2e}

\usepackage[utf8]{inputenc}

\usepackage{lipsum}

\usepackage{color, soul}

\usepackage{graphicx}

\newcommand{\R}{\mathbb{R}}
\newcommand{\E}{\mathbb{E}}
\newcommand{\NN}{\mathcal{N}}

\newcommand{\mt}{text}
\newcommand{\mv}{visual}
\newcommand{\ma}{acoustic}
\newcommand{\mT}{Text}
\newcommand{\mV}{Visual}
\newcommand{\mA}{Acoustic}

\newcommand{\MI}{MI}

\newcommand{\modelname}{MMIM}

\title{Improving Multimodal Fusion with Hierarchical Mutual Information Maximization for Multimodal Sentiment Analysis}

\author{Wei Han$^\dagger$, Hui Chen$^\dagger$, Soujanya Poria$^\dagger$,\\
  $^\dagger$ Singapore University of Technology and Design, Singapore\\
  \texttt{\{wei\_han, hui\_chen\}@mymail.sutd.edu.sg}\\
  \texttt{sporia@sutd.edu.sg}\\
  }

\begin{document}
\maketitle

\begin{abstract}
In multimodal sentiment analysis (MSA), the performance of a model highly depends on the quality of synthesized embeddings.
These embeddings are generated from the upstream process called multimodal fusion, which aims to extract and combine the input unimodal raw data to produce a richer multimodal representation.
Previous work either back-propagates the task loss or manipulates the geometric property of feature spaces to produce favorable fusion results, which neglects the preservation of critical task-related information that flows from input to the fusion results.
In this work, we 
propose a framework named MultiModal InfoMax (\modelname), which hierarchically maximizes the Mutual Information (MI) 
in unimodal input pairs (inter-modality) and between multimodal fusion result and unimodal input
in order to maintain task-related information through multimodal fusion.
The framework is jointly trained with the main task (MSA) to improve the performance of the downstream MSA task.
To address the intractable issue of MI bounds, we further formulate a set of computationally simple parametric and non-parametric methods to approximate their truth value.
Experimental results on the two widely used datasets demonstrate the efficacy of our approach. The implementation of this work is publicly available at \url{https://github.com/declare-lab/Multimodal-Infomax}.
\end{abstract}
\section{Introduction}
With the unprecedented advances in social media in recent years and the availability of smartphones with high-quality cameras, we witness an explosive boost of multimodal data, such as movies, short-form videos, etc.
In real life, multimodal data usually consists of three channels: \mv~(image), \ma~(voice), and transcribed text. 
Many of them often express sort of sentiment, which is a long-term disposition evoked when a person encounters a specific topic, person or entity~\citep{deonna2012emotions,poria2020beneath}.
Mining and understanding these emotional elements from multimodal data, namely multimodal sentiment analysis (MSA), has become a hot research topic because of numerous appealing applications, such as obtaining overall product feedback from customers or gauging polling intentions from potential voters~\citep{melville2009sentiment}. 
Generally, different modalities in the same data segment are often complementary to each other, providing extra cues for semantic and emotional disambiguation~\citep{ngiam2011multimodal}. 
The crucial part for MSA is multimodal fusion, in which a model aims to extract and integrate information from all input modalities to understand the sentiment behind the seen data.
Existing methods to learn unified representations are grouped
in two categories: through loss back-propagation or geometric manipulation in the feature spaces.
The former only tunes the parameters based on back-propagated gradients from the task loss~\citep{zadeh2017tensor, tsai2019multimodal, ghosal-etal-2019-dialoguegcn}, reconstruction loss~\citep{mai2020modality}, or auxiliary task loss~\citep{chen2017multimodal,yu2021learning}.
The latter additionally rectifies the spatial orientation of unimodal or multimodal representations by matrix decomposition~\citep{liu2018efficient} or Euclidean measure optimization~\citep{sun2020learning, hazarika2020misa}.
Although having gained excellent results in MSA tasks, these methods are limited to the lack of control in the information flow 
that starts from raw inputs till the fusion embeddings, which may risk losing practical information and introducing unexpected noise carried by each modality~\cite{tsai2020multimodal}. 
To alleviate this issue, different from previous work, we leverage the functionality of mutual information (MI), a concept from the subject of information theory.
MI measures the dependencies between paired multi-dimensional variables.
Maximizing MI has been demonstrated efficacious in removing redundant information irrelevant
to the downstream task and capturing invariant trends or messages across time or different domains~\citep{poole2019variational}, and has been shown remarkable success in the field of representation learning~\cite{hjelm2018learning,velivckovic2018deep}. 
Based on these experience, we propose MultiModal InfoMax (\modelname), a framework that hierarchically maximizes the mutual information in multimodal fusion.
Specifically, we enhance two types of mutual information in representation pairs: between unimodal representations and between fusion results and their low-level unimodal representations.
Due to the intractability of mutual information~\citep{belghazi2018mutual}, researchers always boost MI lower bound instead for this purpose.
However, we find it is still difficult to figure out some terms in the expressions of these lower bounds in our formulation.
Hence for convenient and accurate estimation of these terms, we propose a hybrid approach composed of parametric and non-parametric parts based on data and model characteristics. 
The parametric part refers to neural network-based methods, and in the non-parametric part we exploit a Gaussian Mixture Model (GMM) with learning-free parameter estimation.
Our contributions can be summarized as follows:
\begin{enumerate}
    \item We propose a hierarchical MI maximization framework for multimodal sentiment analysis. MI maximization occurs at the input level and fusion level to reduce the loss of valuable task-related information. 
    To our best knowledge, this is the first attempt to bridge MI and MSA.
    \item We formulate the computation details in our framework to solve the intractability problem. 
    The formulation includes parametric learning and non-parametric GMM with stable and smooth parameter estimation.
    \item We conduct comprehensive experiments on two publicly available datasets and gain superior or comparable results to the state-of-the-art models.
\end{enumerate}

\section{Related Work}
In this section, we briefly overview some related work in multimodal sentiment analysis and mutual information estimation and application.

\subsection{Multimodal Sentiment Analysis (MSA)}
MSA is an NLP task that collects and tackles data from multiple resources such as \ma, \mv, and textual information to comprehend varied human emotions~\citep{morency2011towards}.
Early fusion models adopted simple network architectures, such as RNN based models~\citep{wollmer2013youtube,chen2017multimodal} that capture temporal dependencies from low-level multimodal inputs, SAL-CNN~\citep{wang2017select} which designed a select-additive learning procedure to improve the generalizability of trained neural networks, etc. 
Meanwhile, there were many trials to combine geometric measures as accessory learning goals into deep learning frameworks.
For instance, \citet{hazarika2018cascade, sun2020learning} optimized the deep canonical correlation between modality representations for fusion and then passed the fusion result to downstream tasks.
More recently, formulations influenced by novel machine learning topics have emerged constantly:
\citet{akhtar2019multi} presented a deep multi-task learning framework to jointly learn sentiment polarity and emotional intensity in a multimodal background.
\citet{pham2019found} proposed a method that cyclically translates between modalities to learn robust joint representations for sentiment analysis. 
\citet{tsai2020multimodal} proposed a routing procedure that dynamically adjusts weights among modalities to provide interpretability for multimodal fusion.
Motivated by advances in the field of domain separation, \citet{hazarika2020misa} projected modality features into private and common feature spaces to capture exclusive and shared characteristics across different modalities.
\citet{yu2021learning} designed a multi-label training scheme that generates extra unimodal labels for each modality and concurrently trained with the main task.
\par
In this work, we build up a hierarchical MI-maximization guided model to improve the fusion outcome as well as the performance in the downstream MSA task, where MI maximization is realized not only between unimodal representations but also between fusion embeddings and unimodal representations.

\subsection{Mutual Information in Deep Learning}
Mutual information (MI) is a concept from information theory that estimates the relationship between pairs of variables. It is a reparameterization-invariant measure of dependency~\citep{tishby2015deep} defined as:
\begin{equation}
    I(X;Y)=\E_{p(x,y)}\left[\log\frac{p(x,y)}{p(x)p(y)}\right]
\end{equation}
\citet{alemi2016deep}~first combined MI-related optimization into deep learning models.
From then on, numerous works studied and demonstrated the benefit of the MI-maximization  principle~\citep{bachman2019learning,he2020momentum, amjad2019learning}.
However, since direct MI estimation in high-dimensional spaces is nearly impossible, many works attempted to approximate the true value with variational bounds~\citep{belghazi2018mutual, cheng2020club, poole2019variational}.
\par
In our work, we apply MI lower bounds at both the input level and fusion level and formulate or reformulate estimation methods for these bounds based on data characteristics and mathematical properties of the terms to be estimated.

\section{Method}
\subsection{Problem Definition}
In MSA tasks, the input to a model is unimodal raw sequences
$X_m\in \R^{l_m\times d_m}$ drawn from the same video fragment, where $l_m$ is the sequence length and $d_m$ is the representation vector dimension of modality $m$, respectively. 
Particularly, in this paper we have $m\in \{t, v, a\}$, where $t,v,a$ denote the three types of modalities---\mt, \mv~and~\ma~that we obtained from the datasets.
The goal for the designed model is to extract and integrate task-related information from these input vectors to form a unified representation and then utilize that to make accurate predictions about a truth value $y$ that reflects the sentiment strength.

\begin{figure*}
    \centering
    \includegraphics [trim=0cm 0cm 0cm 0cm, width=\textwidth]{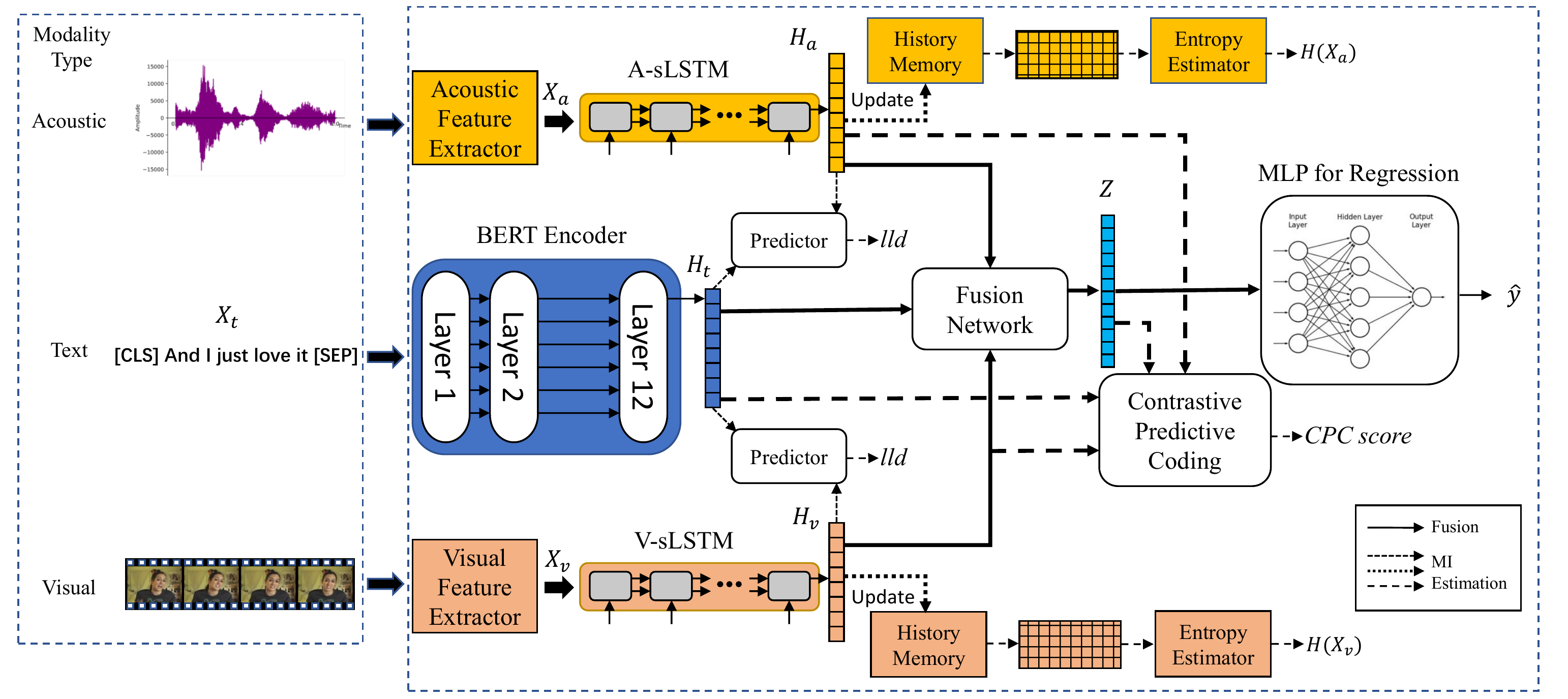}
    \caption{The overall architecture of the MMIM model.}
    \label{MMIM}
\end{figure*}

\subsection{Overall Architecture}
As shown in Figure \ref{MMIM}, our model firstly processes raw input into numerical sequential vectors with feature extractor (firmware for \mv\ and \ma\ with no parameters to train) and tokenizer (for text). 
Then we encode them into individual unit-length representations. 
The model then works in two collaborative parts---fusion and MI maximization, marked by solid and dash lines in Figure \ref{MMIM} respectively.
In the fusion part, a fusion network $F$ of stacked linear-activation layers transforms the unimodal representations into the fusion result $Z$, which is then passed through a regression multilayer perceptron (MLP) for final predictions. 
In the MI part, the MI lower bounds at two levels---input level and fusion level are estimated and boosted.
The two parts work concurrently to produce task and MI-related losses for back-propagation, through which the model learns to infuse the task-related information into fusion results as well as improve the accuracy
of predictions in the main task.

\subsection{Modality Encoding}
We firstly encode the multimodal sequential input $X_m$ into unit-length representations $h_m$. 
Specifically, we use BERT~\citep{devlin2019bert}~to encode an input sentence and extract the head embedding from the last layer's output as $h_t$. 
For \mv~and~\ma, following previous works~\citep{hazarika2020misa, yu2021learning}, we employ two modality-specific unidirectional LSTMs~\citep{hochreiter1997long}~to capture the temporal features of these modalities:
\begin{equation}
\begin{gathered}
    h_t = \textrm{BERT}\left(X_t;\theta_t^{BERT}\right) \\
    h_m = \textrm{sLSTM}\left(X_m; \theta_m^{LSTM}\right) \; m\in\{v,a\}
\end{gathered}
\label{encoding}
\end{equation}

\subsection{Inter-modality MI Maximization}
For a modality representation pair $X, Y$ that comes from a single video clip, although they seem to be independent sequences, there is a certain correlation between them~\cite{arandjelovic2017look}.
Formally, suppose we have a collection of videos $V$ and assume that their prior distributions are known. 
Then the prior distribution of $X$ and $Y$ can be decomposed by the sampling process in $V$ as $P(X) = \int_V P(X|V)P(V)$ and $P(Y) = \int_V P(Y|V)P(V)$, as well as their joint distribution $P(X,Y)=\int_V P(X,Y|V)P(V)$. Unless $P(X,Y|V)=P(X|V)P(Y|V)$, i.e., $X$ and $Y$ are conditionally independent from $V$, the~\MI~is never trivially 0. 
\par
Since the analysis above, we hope that through prompting~\MI~between multimodal input we can filter out modality-specific random noise that is irrelevant to our task and keep modality-invariant contents that span all modalities as much as possible.
As stated before, we boost a tractable lower bound instead of computing MI directly for this purpose.
We exploit an accurate and straightforward \MI~lower bound introduced in~\citet{agakov2004algorithm}.
It approximates the truth conditional distribution $p(y|x)$ with a variational counterpart $q(y|x)$:
\begin{equation}
\begin{aligned}
    I(X;Y) = &\E_{p(x,y)}\left[\log\frac{q(y|x)}{p(y)}\right] + \\ &\E_{p(y)}\left[KL(p(y|x)\Vert q(y|x))\right] \\ 
    \geq &\E_{p(x,y)}\left[\log q(y|x)\right]+H(Y) \\
    \triangleq &I_{BA}
\end{aligned}
\label{Inter-modality MI}
\end{equation}
where $H(Y)$ is the differential entropy of $Y$. This lower bound is tight, i.e., there is no gap between the bound and truth value, when $q(y|x)=p(y|x)$.
In our implementation, we optimize the bounds for two modality pairs--- (text, visual) and (text, acoustic). 
In each pair, we treat text as $X$ and the other modality as $Y$ in~\eqref{Inter-modality MI}. 
We do so because 1) Since we have to train a predictor $q(y|x)$ to approximate $p(y|x)$, prediction from higher-dimensional vectors $h_t\in \R^{d_t}$ ($d_t$ =768) to lower ones $h_v\in \R^{d_v}$ and $h_a\in \R^{d_a}$ ($d_v,d_a$ < 50) converges faster with higher accuracy; 2) many previous works~\citep{tsai2019multimodal,hazarika2020misa} pointed out that from empirical study the text modality is predominate, which can integrate more task-related features than other modalities in this step.
Additionally, we examine the efficacy of the design choice in the ablation study part.
Following \citet{cheng2020club}, we formulate $q(\bm{y}|\bm{x})$ as a multivariate Gaussian distributions $q_{\theta}(\bm{y}|\bm{x})=\NN (\bm{y}|\bm{\mu}_{\theta_1}(\bm{x}),\bm{\sigma}^2_{\theta_2}(\bm{x})\bm{I})$, with two neural networks parameterized by $\theta_1$ and $\theta_2$ to predict the mean and variance, respectively. 
The loss function for likelihood maximization is:
\begin{equation} \label{lld}
    \mathcal{L}_{lld} = -\frac{1}{N}\sum_{tv,ta}\sum_{i=1}^N \log q(y_i|x_i)
\end{equation}
where $N$ is the batch size in training, $tv, ta$ means summing the likelihood of two predictors.

For the entropy term $H(Y)$, we solve its computation with the Gaussian Mixture Model (GMM), a commonly utilized approach for unknown distribution approximation that can facilitate distribution-based estimation~\citep{nilsson2002gaussian,kerroum2010textural}.
GMM builds up multiple Gaussian distributions for different property classes.
We choose the sentiment polarity (non-negative/negative), which is a natural property in the datasets, as the classification criterion, which can also balance the trade-off between estimation accuracy (requires more classes) and computational cost (requires fewer classses).
We build up two normal distributions $\mathcal{N}_{pos}(\bm{\mu}_1,\bm{\Sigma}_1)$ and $\mathcal{N}_{neg}(\bm{\mu}_2,\bm{\Sigma}_2)$ for each class, where $\bm{\mu}$ is the mean vector and $\bm{\Sigma}$ is the covariance matrix. 
The parameters are estimated via the maximum likelihood method on a sufficiently large sampling batch $\mathcal{D}_s \subset \mathcal{D}_{train}$:
\begin{equation}
\begin{gathered} 
    \hat{\bm{\mu}}_c = \frac{1}{N_c}\sum_{i=1}^{N_c}h^i_c \\
    \hat{\bm{\Sigma}}_c = \frac{1}{N_c}\sum_{i=1}^{N_c}h^i_c \odot h^i_c - \hat{\bm{\mu}}_c^T\hat{\bm{\mu}}_c
\end{gathered}
\label{Gest1}
\end{equation}
where $c \in \{pos,neg\}$ represents the polarity class that the sample belongs to, $N_c$ is the number of samples in class $c$ and $\odot$ is component-wise multiplication. 
The entropy of a multivariate normal distribution is given by:
\begin{equation} 
    H = \frac{1}{2} \log\left((2\pi e)^k\det(\bm{\Sigma})\right) = \frac{\log(\det(2\pi e\bm{\Sigma}))}{2}
\end{equation}
where $k$ is the dimensionality of the vectors in GMM and $\det(\bm{\Sigma})$ is the determinant of $\bm{\Sigma}$.
Based on the nearly equal frequencies of the two polarity classes in the dataset, we assume the prior probability that one data point $\bm{x}=(x_1,...,x_k)$ belongs to each is equal, i.e., $w_{pos}=p(\bm{x}\in pos)=w_{neg}=p(\bm{x}\in neg)=\frac{1}{2}$. 
Under the assumption that the two sub-distributions are disjoint, from \citet{huber2008entropy} the lower and upper bound of a GMM's entropy are:
\begin{equation}
    \sum_c w_ch_c \leq H(Y) \leq \sum_c w_c(-\log w_c +h_c)
\end{equation}
where $h_c$ is the entropy of the sub-distribution for class $c$.
Taking the lower bound as an approximation, we obtain the entropy term for the MI lower bound:
\begin{equation} \label{entropy estimation}
    H(Y) = \frac{1}{4}\left[\log((\det(\bm{\Sigma}_1)\det(\bm{\Sigma}_2))\right]
\end{equation}
In this formulation, we implicitly assume that the prior probabilities of the two classes are equal. 
We further notice that $H(Y)$ changes every time during each training epoch but at a very slow pace in several continuous steps due to the small gradients and consequently slight fluctuation in parameters.
This fact demands us to update parameters timely to ensure estimation accuracy.
Besides, according to statistical theory, we should increase the batch size ($N_*$) to reduce estimation error, but the maximum batch size is restricted to the GPU's capacity.
Considering the situation above, we indirectly enlarge $\mathcal{D}_s$ by encompassing the data from the nearest history.
In implmentation, we store such data in a history data memory.
The loss function for MI lower bound maximization in this level is given by:
\begin{equation} \label{BA Loss}
    \mathcal{L}_{BA} = -I_{BA}^{t,v} - I_{BA}^{t,a}
\end{equation} \label{LBA}
\label{MI1}

\subsection{MI Maximization in the Fusion Level} \label{CPC}
To enforce the intermediate fusion results to capture modality-invariant cues among modalities, we repeat MI maximization between fusion results and input modalities.
The optimization target is the fusion network $F$ that produces  fusion results $Z=F(X_t, X_v, X_a)$.
Since we already have a generation path from $X_m$ to $Z$, we expect an opposite path, i.e. to constructs $X_m, m\in\{t,v,a\}$ from $Z$. Inspired by but different from~\citet{oord2018representation}, we use a score function that acts on the normalized prediction and truth vectors to gauge their correlation:
\begin{equation}
\begin{gathered}
\overline{G_\phi(Z)} = \frac{G_\phi(Z)}{\Vert G_\phi(Z) \Vert_2}, \quad \overline{h_m} = \frac{h_m}{\Vert h_m \Vert}_2 \\ 
s(h_m, Z)=\exp{\left(\overline{h_m} \left(\overline{G_\phi(Z)}\right)^T\right)}
\end{gathered}
\label{Score Function}
\end{equation}
where $G_\phi$ is a neural network with parameters $\phi$ that generates a prediction of $h_m$ from $Z$, $\Vert \cdot \Vert_2$ is the Euclidean norm, by dividing which we obtain unit-length vectors.
Because we find the model intends to stretch both vectors to maximize the score in~\eqref{Score Function} without this normalization.
Then same as what \citet{oord2018representation} did, we incorporate this score function into the Noise-Contrastive Estimation framework~\citep{gutmann2010noise}~by treating all other representations of that modality in the same batch $\tilde{\mathbf{H}}_m^i = \mathbf{H}_m \setminus \{h_m^i\}$~as negative samples:
\begin{equation}\label{LN}
    \mathcal{L}_{\mathbf{N}}(Z, \mathbf{H}_m) = -\E_{\mathbf{H}}\left[\log \frac{s(Z,h_m^i)}{\sum_{h_m^j \in \mathbf{H}_m}s(Z, h_m^j)} \right]
\end{equation}
Here is a short explanation of the rationality of such formulation.
Contrastive Predictive Coding (CPC) scores the MI between context and future elements ``across the time horizon'' to keep the portion of ``slow features'' that span many time steps~\citep{oord2018representation}.
Similarly, in our model, we ask the fusion result $Z$ to reversely predict representations ``across modalities'' so that more modality-invariant information can be passed to $Z$. 
Besides, by aligning the prediction to each modality we enable the model to decide how much information it should receive from each modality. 
This insight will be further discussed with experimental evidence in Section~\ref{Case Study}.
The loss function for this level is given by:
\begin{equation}\label{LCPC}
    \mathcal{L}_{CPC} = \mathcal{L}_{\mathbf{N}}^{z,v}+\mathcal{L}_{\mathbf{N}}^{z,a} + \mathcal{L}_{\mathbf{N}}^{z,t}
\end{equation}
\begin{algorithm}[ht!]
\small
\SetAlgoLined
\KwIn{Dataset $\mathcal{D} = \{(X_t,X_v, X_a),Y\}$, $\alpha$, $\beta$, learning rate $\eta_{lld}, \eta_{main}$, embedding history memory $M$}
\KwOut{Prediction $\hat{y}$}
\For{\upshape each training epoch}{
\itshape Stage 1: Conditional Likelihood Maximization \\
\For{\upshape minibatch $\mathcal{B}=\{(X_t^i,X_v^i, X_a^i)\}_{i=1}^N$ sampled from $\mathcal{D}_{sub} \subseteq \mathcal{D}$}{
    \upshape
    Encode $X_m^i$ to $h_m^i$ as \eqref{encoding} \\
    Compute $\mathcal{L}_{lld}$ as \eqref{lld} \\
    Update parameters of predictor $q$:
    $\theta_q \leftarrow \theta_q - \eta_{lld}\nabla_{\theta}\mathcal{L}_{lld}$
 }
Stage 2: MI-maximization Joint Training: \\
 \For{\upshape minibatch $\mathcal{B}=\{(X_t^i,X_v^i, X_a^i)\}_{i=1}^N$ sampled from $\mathcal{D}$}{
    \upshape
    Encode $X_m^i$ to $h_m^i$ as \eqref{encoding} \\
    Estimate the mean vectors and co-variance matrices in GMM model with $M$ as \eqref{Gest1}\\
    Update history $M$:
    $M \leftarrow M \setminus \{Oldest\;Hidden\;Batch\}$
    $M \leftarrow  M \cup \{h_m^i\}_{i=1}^{|B|}, m\in\{v,a\}$ \\
    Compute $\mathcal{L}_{BA}$ as~\eqref{Inter-modality MI}, \eqref{entropy estimation}, \eqref{BA Loss}\\
    Produce fusion results $Z_i = F(X_t^i, X_v^i, X_a^i)$ and predictions $\hat{y}$\\
    Compute $\mathcal{L}_{N}$, $\mathcal{L}_{CPC}$ as \eqref{Score Function}, \eqref{LN}, \eqref{LCPC} \\
    Compute $\mathcal{L}_{main}$ as \eqref{main loss} \\
    Update all parameters in the model except $q$:
    $\theta_k \leftarrow \theta_k - \eta_k\nabla_{\theta}\mathcal{L}_{main}$}
 }
\caption{\small MultiModal Mutual Information Maximization (MM-MIM)}
\label{alg:1}
\end{algorithm}

\subsection{Training}
The training process consists of two stages in each iteration:
In the first stage, we approximate $p(y|x)$ with $q(y|x)$ by minimizing the negative log-likelihood for inter-modality predictors with the loss in \eqref{lld}.
In the second stage, hierarchical MI lower bounds in previous subsections are added to the main loss as auxiliary losses.
After obtaining the final prediction $\hat{y}$, along with the truth value $y$, we have the task loss: 
\begin{equation}
    \mathcal{L}_{task} = \textrm{MAE}(\hat{y}, y)
\end{equation}
where MAE stands for mean absolute error loss, which is a common practice in regression tasks. 
Finally, we calculate the weighted sum of all these losses to obtain the main loss for this stage:
\begin{equation} \label{main loss}
    \mathcal{L}_{main} = \mathcal{L}_{task} + \alpha \mathcal{L}_{CPC} + \beta \mathcal{L}_{BA}
\end{equation}
where $\alpha, \beta$ are hyper-parameters that control the impact of MI maximization.
We summarize the training algorithm in Algorithm~\ref{alg:1}.

\section{Experiments}
In this section, we present some experimental details, including datasets, baselines, feature extraction tool kits, and results.

\subsection{Datasets and Metrics}
We conduct experiments on two publicly available academic datasets in MSA research:  CMU-MOSI~\citep{zadeh2016multimodal} and CMU-MOSEI~\citep{zadeh2018multimodal}. 
CMU-MOSI contains 2199 utterance video segments sliced from 93 videos in which 89 distinct narrators are sharing opinions on interesting topics.
Each segment is manually annotated with a sentiment value ranged from -3 to +3, indicating the polarity (by positive/negative) and relative strength (by absolute value) of expressed sentiment. 
CMU-MOSEI dataset upgrades CMU-MOSI by expanding the size of the dataset. 
It consists of 23,454 movie review video clips from YouTube. 
Its labeling style is the same as CMU-MOSI.
We provide the split specifications of the two datasets in Table~\ref{Dataset Specification}.
\begin{table}
    \centering
    \begin{tabular}{ccc}
        \hline
        Split & CMU-MOSI & CMU-MOSEI \\
        \hline
         Train & 1284 & 16326 \\
         Validation & 229 & 1871 \\
         Test & 686 & 4659 \\
         \hdashline
         All & 2199 & 22856 \\
         \hline
    \end{tabular}
    \caption{Dataset split.}
    \label{Dataset Specification}
\end{table}
\par
We use the same metric set that has been consistently presented and compared before: mean absolute error (MAE), which is the average mean absolute difference value between predicted values and truth values, Pearson correlation (Corr) that measures the degree of prediction skew, seven-class classification accuracy (Acc-7) indicating the proportion of predictions that correctly fall into the same interval of seven intervals between -3 and +3 as the corresponding truths, binary classification accuracy (Acc-2) and F1 score computed for positive/negative and non-negative/negative classification results.

\begin{table*}[ht]
    \centering
     \resizebox{\linewidth}{!}{
    \begin{tabular}{l|c c c c c|c c c c c}
        \toprule
        \multirow{2}{6em}{$\textrm{models}^{\diamondsuit}$} & \multicolumn{5}{c|}{CMU-MOSI} & \multicolumn{5}{c}{CMU-MOSEI\rule[-1.5ex]{0ex}{0ex}} \\
        ~ & MAE & Corr & Acc-7 & Acc-2 & F1 & MAE & Corr & Acc-7 & Acc-2 & F1 \\
        \midrule
        $\textrm{TFN}^{\dag}$ & 0.901 & 0.698 & 34.9 & - /80.8 & - /80.7 & 0.593 & 0.700 & 50.2 & - /82.5 & - /82.1 \\
        $\textrm{LMF}^{\dag}$ & 0.917 & 0.695 & 33.2 & - /82.5 & - /82.4 & 0.623 & 0.677 & 48.0 & - /82.0 & - /82.1 \\
        $\textrm{MFM}^{\dag}$ & 0.877 & 0.706 & 35.4 & - /81.7 & - /81.6 & 0.568 & 0.717 & 51.3 & - /84.4 & - /84.3  \\
        $\textrm{ICCN}^{\dag}$ & 0.862 & 0.714 & 39.0 & - /83.0 & - /83.0  & 0.565 & 0.713 & 51.6 & - /84.2 & - /84.2 \\
        $\textrm{MulT}^{\ddag}$ & 0.861 & 0.711 & - & 81.5/84.1 & 80.6/83.9 & 0.580 & 0.703 & - & - /82.5 & - /82.3 \\
        $\textrm{MISA}^{\ddag}$ & 0.804 & 0.764 & - & 80.79/82.10 & 80.77/82.03 & 0.568 & 0.724 & - & 82.59/84.23 & 82.67/83.97 \\
        $\textrm{MAG-BERT}^{\ddag}$ & 0.731 & 0.789 & - & 82.5/84.3 & 82.6/84.3 & 0.539 & 0.753 & - & 83.8/85.2 & 83.7/85.1 \\
        $\textrm{Self-MM}^{\ddag}$ & 0.713 & 0.798 & - & 84.00/85.98 & 84.42/85.95 & 0.530 & 0.765 & - & 82.81/85.17 & 82.53/85.30 \\
        \midrule
        $\textrm{MAG-BERT}^{\ast}$ & 0.727 & 0.781 & 43.62 & 82.37/84.43 & 82.50/84.61 & 0.543 & 0.755 & 52.67 & 82.51/84.82 & 82.77/84.71 \\
        $\textrm{Self-MM}^{\ast}$ & 0.712 & 0.795 & 45.79 & 82.54/84.77 & 82.68/84.91 & 0.529 & 0.767 & 53.46 & \textbf{82.68}/84.96 & \textbf{82.95}/84.93 \\
        \modelname & \textbf{0.700}$^{\natural}$ & \textbf{0.800}$^{\natural}$ & \textbf{46.65}$^{\natural}$ & \textbf{84.14}$^{\natural}$/\textbf{86.06}$^{\natural}$ & \textbf{84.00$^{\natural}$/85.98$^{\natural}$} & \textbf{0.526} & \textbf{0.772} & \textbf{54.24$^{\natural}$} & 82.24/\textbf{85.97$^{\natural}$} & 82.66/\textbf{85.94$^{\natural}$} \\
        \bottomrule
    \end{tabular}
    }
    \caption{Results on CMU-MOSI and CMU-MOSEI; $\diamondsuit$: all models use BERT as the text encoder;  $\dag$: from~\citet{hazarika2020misa}; $\ddag$: from~\citet{yu2021learning}; $\ast$: reproduced from open-source code with hyper-parameters provided in original papers. For Acc-2 and F1, we have two sets of non-negative/negative (left) and positive/negative (right) evaluation results. Best results are marked in bold and $\natural$ means the corresponding result is significantly better than SOTA with p-value < 0.05 based on paired t-test.}
    \label{Experiment Results}
\end{table*}

\subsection{Baselines}
To inspect the relative performance of~\modelname, we compare our model with many baselines. 
We consider pure learning based models, such as TFN~\citep{zadeh2017tensor}, LMF~~\citep{liu2018efficient}, MFM~\citep{tsai2019learning} and MulT~\citep{tsai2019multimodal}, as well as approaches involving feature space manipulation like ICCN~\citep{sun2020learning} and MISA~\citep{hazarika2020misa}.
We also compare our model with more recent and competitive baselines, including BERT-based model---MAG-BERT~\citep{rahman-etal-2020-integrating} and Self-MM~\citep{yu2021learning}, which works with multi-task learning and is the SOTA method. Some of the baselines are available at \url{https://github.com/declare-lab/multimodal-deep-learning}.

The baselines are listed below:

\textbf{TFN}~\citep{zadeh2017tensor}: 
Tensor Fusion Network disentangles unimodal into tensors by three-fold Cartesian product. 
Then it computes the outer product of these tensors as fusion results.

\textbf{LMF}~\citep{liu2018efficient}: Low-rank Multimodal Fusion decomposes stacked high-order tensors into many low rank factors then performs efficient fusion based on these factors.

\textbf{MFM}~\citep{tsai2019learning}: Multimodal Factorization Model concatenates a inference network and a generative network with intermediate modality-specific factors, to facilitate the fusion process with reconstruction and discrimination losses.

\textbf{MulT}~\citep{tsai2019multimodal}: Multimodal Transformer constructs an architecture  unimodal and crossmodal transformer networks and complete fusion process by attention.

\textbf{ICCN}~\citep{sun2020learning}: Interaction Canonical Correlation Network minimizes canonical loss between modality representation pairs to ameliorate fusion outcome.

\textbf{MISA}~\citep{hazarika2020misa}: Modality-Invariant and -Specific Representations projects features into separate two spaces with special limitations. Fusion is then accomplished on these features.


\textbf{MAG-BERT}~\citep{rahman-etal-2020-integrating}: Multimodal Adaptation Gate for BERT designs an alignment gate and insert that into vanilla BERT model to refine the fusion process.

\textbf{SELF-MM}~\citep{yu2021learning}: Self-supervised Multi-Task Learning assigns each modality a unimodal training task with automatically generated labels, which aims to adjust the gradient back-propagation.

\subsection{Basic Settings and Results}
\paragraph{Experimental Settings.} 
We use unaligned raw data in all experiments as in~\citet{yu2021learning}.
For \mv~and~\ma, we use COVAREP~\citep{degottex2014covarep} and P2FA~\citep{yuan2008speaker}, which both are prevalent tool kits for feature extraction and have been regularly employed before.
We trained our model on a single RTX 2080Ti GPU and ran a grid search for the best set of hyper-parameters. 
The details are provided in the supplementary file.

\paragraph{Hyperparameter Setting}
We perform a grid-search for the best set of hyper-parameters: batch size in \{32, 64\}, $\eta_{lld}$ in \{1e-3,5e-3\}, $\eta_{main}$ in \{5e-4, 1e-3, 5e-3\}, $\alpha$, $\beta$ in \{0.05, 0.1, 0.3\}, hidden dim in \{32, 64\}, memory size in \{1, 2, 3\} batches, gradient clipping value is fixed at 5.0, learning rate for BERT fine-tuning is 5e-5, BERT embedding size is 768 and fusion vector size is 128. The hyperparameters are given in Table \ref{tab:hypers}.

\begin{table}[ht!]
    \centering
    \resizebox{\linewidth}{!}{
    \begin{tabular}{lcc}
        \hline
        Item & CMU-MOSI & CMU-MOSEI \\
        \hline
         batch size & 32 & 64 \\
         learning rate $\eta_{lld}$ & 5e-3 & 1e-3 \\
         learning rate $\eta_{main}$ & 1e-3 & 5e-4 \\
         $\alpha$ & 0.3 & 0.1 \\
         $\beta$ & 0.1 & 0.05 \\
         V-LSTM hidden dim & 32 & 64 \\
         A-LSTM hidden dim & 32 & 16 \\
         memory size & 32 (1 batch) & 64 (1 batch) \\
         gradient clip & 5.0 & 5.0 \\
         \hline
    \end{tabular}
    }
    \caption{Hyperparameters for best performance.}
    \label{tab:hypers}
\end{table}

\paragraph{Summary of the Results.}
In accord with previous work, we ran our model five times under the same hyper-parameter settings and report the average performance in Table~\ref{Experiment Results}.
We find that \modelname~yields better or comparable results to many baseline methods. To elaborate, our model significantly outperforms SOTA in all metrics on CMU-MOSI and in (non-0) Acc-7, (non-0) Acc-2, F1 score on CMU-MOSEI. 
For other metrics, \modelname~achieves very closed performance (<0.5\%) to SOTA.
These outcomes preliminarily demonstrate the efficacy of our method in MSA tasks.

\begin{table}[th]
    \centering
    \resizebox{\linewidth}{!}{
    \begin{tabular}{l|c c c c c}
        \toprule
        Description & MAE & Corr & Acc-7 & Acc-2 & F1 \\
        \midrule
        \modelname\ & \textbf{0.526} & \textbf{0.772} & \textbf{54.24} & \textbf{82.24}/\textbf{85.97} & \textbf{82.66}/\textbf{85.94} \\
        Inter-modality MI \\
        \quad $I_{BA}^{t,v}$ & 0.533 & 0.763 & 53.80 & 80.87/85.08 & 81.37/85.06 \\
        \quad $I_{BA}^{t,a}$ & 0.538 & 0.767 & 53.31 & 80.26/82.73 & 80.81/82.00 \\
        \quad $I_{BA}^{v,a}$ & 0.545 & 0.753 & 53.85 & 80.40/85.05 & 80.85/84.95 \\
        \quad $I_{BA}^{t,a}+I_{BA}^{v,a}$ & 0.536 & 0.764 & 53.53 & 79.40/85.39 & 80.12/85.47 \\
        \quad $I_{BA}^{t,v}+I_{BA}^{v,a}$ & 0.534 & 0.770 & 54.11 & 80.62/85.61 & 81.20/85.64 \\
        \quad $I_{BA}^{t,v}+I_{BA}^{v,a}+I_{BA}^{t,a}$ & 0.527 & \textbf{0.772} & \textbf{54.53} & 80.02/85.42 & 80.64/85.44 \\
        \quad None & 0.541 & 0.752 & 53.57 & 79.60/84.75 & 80.21/84.76 \\
        $\mathcal{L}_{CPC}$ loss \\
        \quad w/o $\mathcal{L}_N^{z,t}$ & 0.535 & 0.768 & 53.66 & 76.46/83.92 & 77.38/84.04 \\
        \quad w/o $\mathcal{L}_N^{z,v}$ & 0.536 & 0.766 & 53.70 & \textbf{82.71}/85.86 & \textbf{82.80}/\textbf{85.97}\\
        \quad w/o $\mathcal{L}_N^{z,a}$ & 0.530 & 0.771 & 53.44 & 80.68/85.78 &  81.18/85.72 \\
        \quad w/o $\mathcal{L}_N^{z,t},\mathcal{L}_N^{z,v},\mathcal{L}_N^{z,a}$ & 0.543 & 0.759 & 53.49 & 78.89/84.37 & 79.57/84.40 \\
        Entropy estimation \\
        \quad w/o history data & NaN & NaN & NaN & NaN/NaN & NaN/NaN  \\
        \quad w/o GMM & 0.533 & 0.768 & 53.4 & 79.57/84.94 & 80.19/84.95  \\
        \bottomrule
    \end{tabular}
     }
    \caption{Ablation study of \modelname~on CMU-MOSEI. $t,v,a,z$ represent text, visual, acoustic and fusion results.}
    \label{Abl Study}
\end{table}
\subsection{Ablation Study} \label{ABLS}
To show the benefits from the proposed loss functions and the corresponding estimation methods in~\modelname, we carried out a series of ablation experiments on CMU-MOSEI. 
The results under different ablation settings are categorized and listed in Table \ref{Abl Study}.
First, we eliminate one or several MI loss terms, for both the inter-modality MI lower bound ($I_{BA}$) and 
CPC loss ($\mathcal{L}^{z,m}_{N}$ where $m \in \{v,a,t\}$), from the total loss. 
We note the manifest performance degradation after removing part of the MI loss, and the results are even worse when removing all terms in one loss than only removing single term, which shows the efficacy of our MI maximization framework.
Besides, by replacing current optimization target pairs in inter-modality MI with single pair or other pair combinations we can not gain better results, which provides experimental evidence for the candidate pair choice in that level. 
Then we test the components for entropy estimation. 
We deactivate the history memory and evaluate $\bm{\mu}$ and $\bm{\Sigma}$ in~\eqref{Gest1} using only the current batch. It is surprising to observe that the training process broke down due to the gradient's ``NaN'' value.
Therefore, the history-based estimation has another advantage of guaranteeing the training stability.
Finally, we substitute the GMM with a unified Gaussian where $\bm{\mu}$ and $\bm{\Sigma}$ are estimated on all samples regardless of their polarity classes.
We spot a clear drop in all metrics, which implies the GMM built on natural class leads to a more accurate estimation for entropy terms.

\section{Further Analysis}
In this section, we dive into our models to explore how it functions in the MSA task. We first visualize all types of losses in the training process, then we analyze some representative cases.

\begin{figure}
    \centering
    \includegraphics[trim=0.1cm 14cm 0.1cm 0cm, width=\columnwidth]{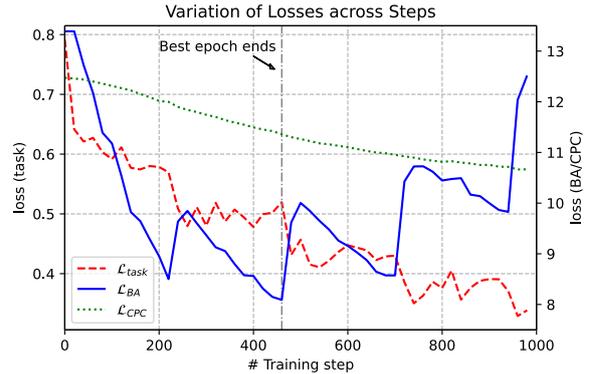}
    \caption{Visualization of loss changing as training proceeds on CMU-MOSEI.}
    \label{Loss vis}
\end{figure}

\begin{table*}[ht]
    \centering
    \resizebox{\linewidth}{!}{
    \begin{tabular}{c  lll  ccc}
        \toprule
          & \multicolumn{1}{c}\mT & \multicolumn1c\mV & \multicolumn1{c}\mA & $s_{zt}/s_{zv}/s_{za}$ & Pred & Truth  \\
        \midrule
      \multirow{2}{*}{(A)} & \multirow{2}*{\shortstack[l]{\it We'll pick it up from here in   \\ \it the next video in this series.}} & \multirow{2}{*}{\shortstack[l]{Smile  }} & \multirow{2}{*}{\shortstack[l]{Slightly rising tone\\Normal volume}} & \multirow{2}{*}{$0.67/\textbf{0.96}/0.43$} & \multirow{2}{*}{$+0.6663$} & \multirow{2}{*}{$+0.6667$}  \\
      & & & & & \\
      \midrule
      \multirow{2}{*}{(B)} & \multirow{2}*{\shortstack[l]{\it I'd probably only give it a two  \\ \it  out of five stars.}} & \multirow{2}{*}{\shortstack[l]{Frown}} & \multirow{2}{*}{\shortstack[l]{Peaceful tone \\ Normal volume}} & \multirow{2}{*}{$\textbf{0.85}/\textbf{0.96}/0.36$} & \multirow{2}{*}{$-1.6642$} & \multirow{2}{*}{$-1.6667$}  \\
      & & & & & \\
        \midrule
      \multirow{2}{*}{(C)} & \multirow{2}*{\shortstack[l]{\it  So these people are commissioned \\ \it  to hunt the animals}} & \multirow{2}{*}{\shortstack[l]{Glance}}  & \multirow{2}{*}{\shortstack[l]{Peaceful \& Narrative}} & \multirow{2}{*}{$0.64/\textbf{0.93}/0.73$} &\multirow{2}{*}{$-0.0009$} & \multirow{2}{*}{$0.0000$}  \\
      & & & & & \\
      \midrule
      \multirow{2}{*}{(D)} & \multirow{2}*{\shortstack[l]{\it I'm sorry, on the scale of one to  \\ \it five I would give this a five.}} & \multirow{2}{*}{\shortstack[l]{Turn head}} &  \multirow{2}{*}{\shortstack[l]{High pitch on ``five''}} & \multirow{2}{*}{$\textbf{0.83}/0.71/0.54$} &  \multirow{2}{*}{$-2.0023$} & \multirow{2}{*}{$+2.6667$}  \\
      & & & & & \\
        \bottomrule
    \end{tabular}
    }
    \caption{Representative examples with their predictions and fusion-modality scores in the case study. High scores ($\geq 0.8$) are highlighted in bold.}
    \label{tab:Case Study}
\end{table*}

\subsection{Tracing the Losses}
To better understand how MI losses work, we visualize the variation of all losses during training in Figure~\ref{Loss vis}. 
The values for plotting are the average losses in a constant interval of every 20 steps.
From the figure, we can see throughout the training process, $\mathcal{L}_{task}$ and $\mathcal{L}_{CPC}$ keep decreasing nearly all the time, while $\mathcal{L}_{BA}$ goes down in an epoch except the beginning of that. 
We also mark the time that the best epoch ends, i.e., the task loss on the validation set reaches the minimum. 
It is notable that $\mathcal{L}_{BA}$ and $\mathcal{L}_{CPC}$ reach a relatively lower level at this time while the task loss on the training set does not. This scenario reveals the crucial role that $\mathcal{L}_{BA}$ and $\mathcal{L}_{CPC}$ play in the training process---they offer supplemental unsupervised gradient rectification to the parameters in their respective back-propagation path and fix up the over-fitting of the task loss.
Besides, because in the experiment settings $\alpha$ and $\beta$ are in the same order and at the end of best epoch $\mathcal{L}_{BA}$ reaches the lowest value, which is synchronized as the validation loss, but $\mathcal{L}_{CPC}$ fails to, we can conclude that  $\mathcal{L}_{BA}$, or MI maximization in the input (lower) level, has a more significant impact on model's performance than $\mathcal{L}_{CPC}$, or MI maximization in the fusion (higher) level.

\subsection{Case Study} \label{Case Study}
We display some predictions and truth values, as well as corresponding input raw data (for \mv~and~\ma~we only illustrate literally) and three CPC scores in Table \ref{tab:Case Study}. 
As described in Section \ref{CPC}, these scores imply how much the fusion results depend on each modality.
It is noted that the scores are all beyond 0.35 in all cases, which demonstrates the fusion results seize a certain amount of domain-invariant features. 
We also observe the different extents that the fusion results depend on each modality. 
In case (A), visual provides the only clue of the truth sentiment, and correspondingly $s_{zv}$ is higher than the other two scores. 
In case (B), the word ``only'' is a piece of additional evidence apart from what visual modality exposes, and we find $s_{zt}$ achieves a higher level than in (A).
For (C), acoustic and visual help infer a neutral sentiment and thus $s_{zv}$ and $s_{za}$ are large than $s_{zt}$.
Therefore, we conclude that the model can intelligently adjust the information that flows from unimodal input into the fusion results consistently with their individual contribution to the final predictions.
However, this mechanism may malfunction in cases like (D). The remark ``I'm sorry'' bewilders the model and meanwhile \mv~and~\ma~remind none. In this circumstance, the model casts attention on text and is misled to a wrong prediction in the opposite direction.


\section{Conclusion}
In this paper, we present \modelname, which hierarchically maximizes the mutual information (MI) in a multimodal fusion pipeline.
The model applies two MI lower bounds for unimodal inputs and the fusion stage, respectively.
To address the intractability of some terms in these lower bounds, we specifically design precise, fast and robust estimation methods to ensure the training can go on normally as well as improve the test outcome.
Then we conduct comprehensive experiments on two datasets followed by the ablation study, the results of which verify the efficacy of our model and the necessity of the MI maximization framework.
We further visualize the losses and display some representative examples to provide a deeper insight into our model.
We believe this work can inspire the creativity in representation learning and multimodal sentiment analysis in the future.

\section*{Acknowledgments}
This project is supported by the AcRF MoE Tier-2 grant titled: ``CSK-NLP: Leveraging Commonsense Knowledge for NLP'', and the SRG grant id: T1SRIS19149 titled ``An Affective Multimodal Dialogue System''.

\bibliography{anthology,custom,citation}
\bibliographystyle{acl_natbib}

\newpage
\appendix

\section{Appendix}
\label{sec:appendix}
\subsection{Implementation details of history memory}
\begin{figure}[ht]
\subfloat[Estimation stage]{%
  \includegraphics[width=\columnwidth, trim=6cm 2cm 5.5cm 2cm, clip]{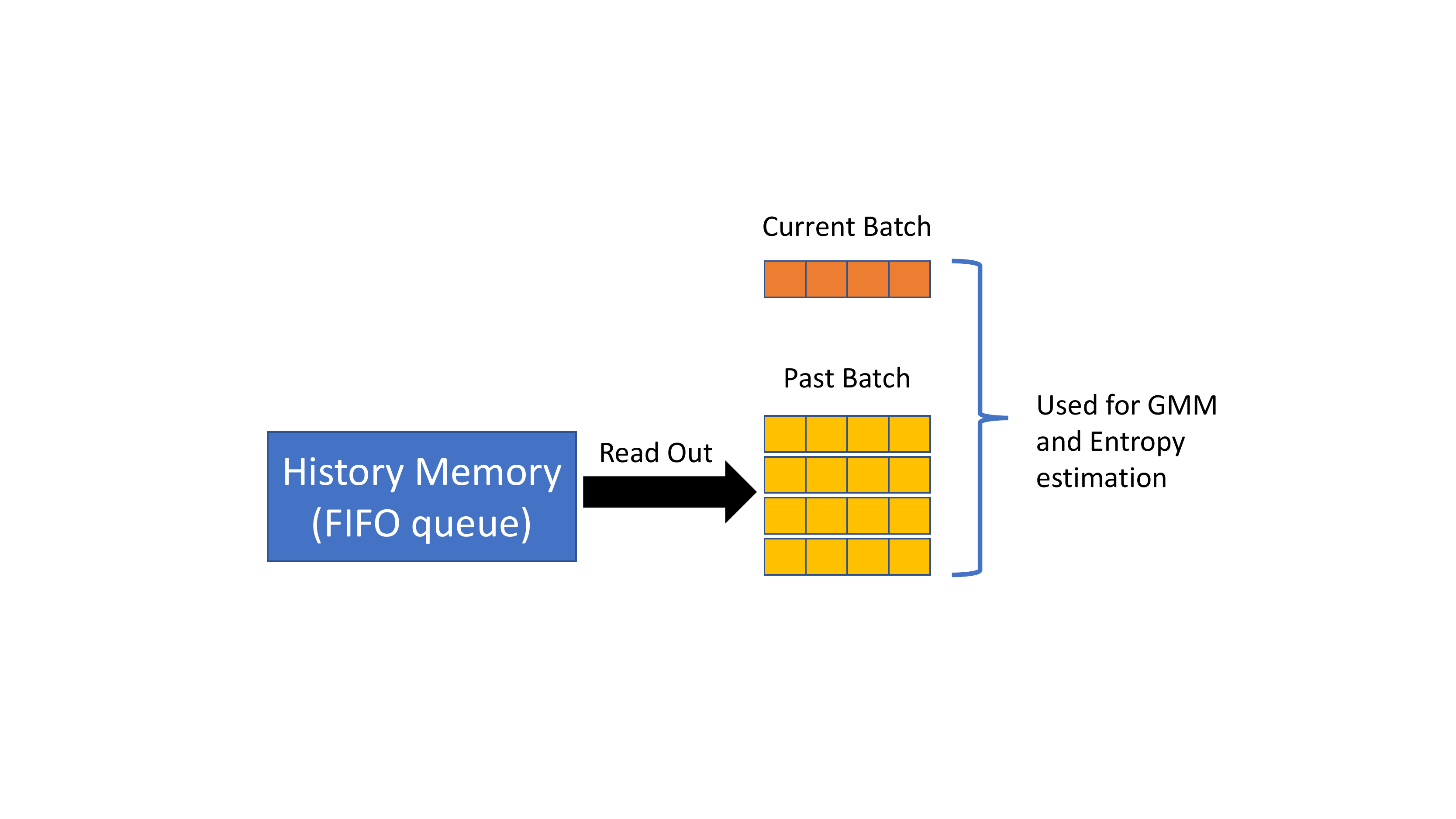}
}

\subfloat[Update stage]{%
 \includegraphics[width=\columnwidth, trim=3cm 4cm 10cm 0cm, clip]{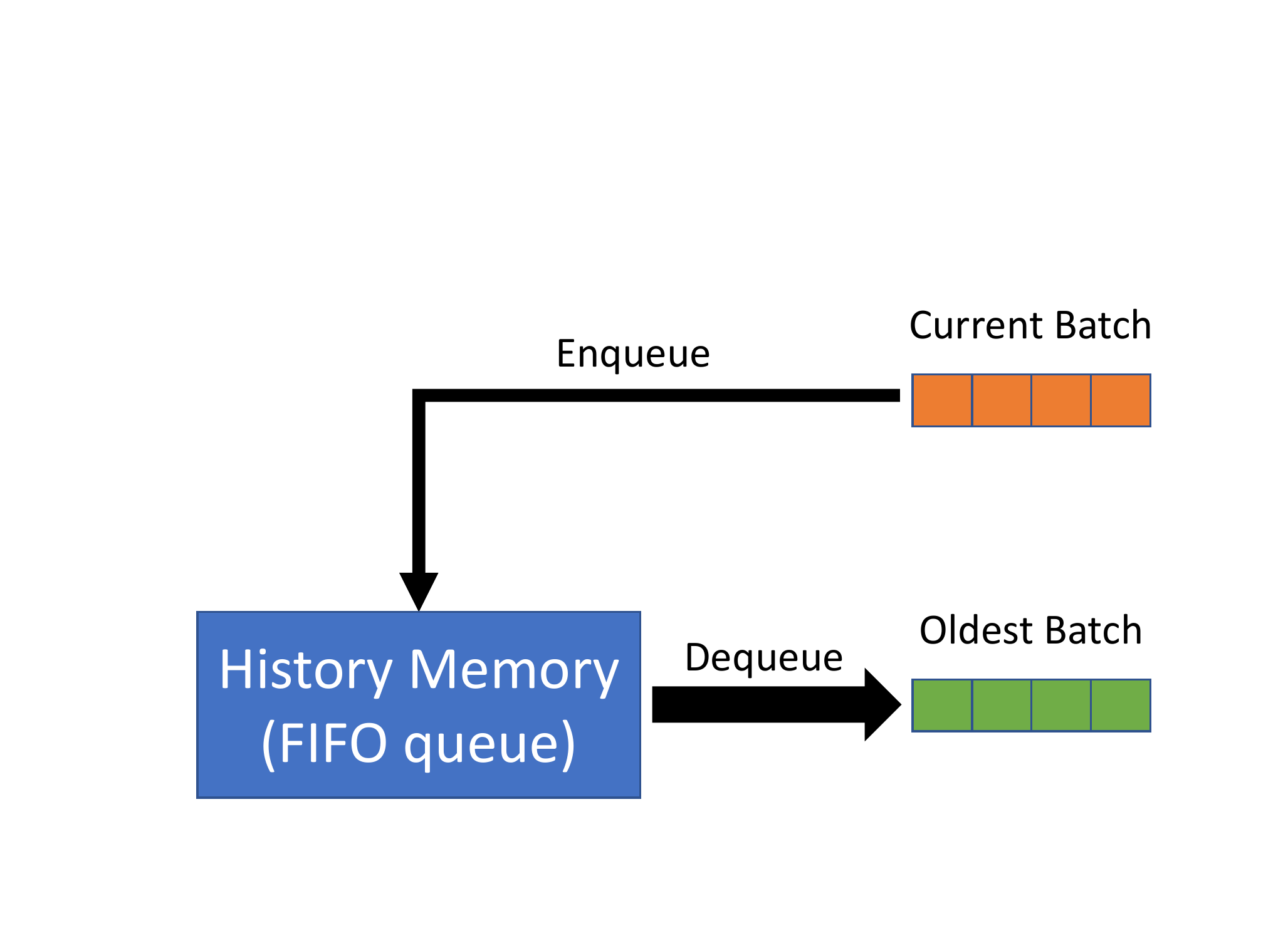}
}
\caption{The workflow of a history embedding memory}
\label{workflow}
\end{figure}

The workflow of the history embedding memory comprises two stages, as shown in Figure \ref{workflow}. In the estimation stage, parameters of the GMM model is estimated using both history embeddings readout from history memory and current batch input, as shown in (a).
Then in the update stage, the oldest batch of data is driven out of the memory to leave space for new data, as described in (b). 
The memory is implemented as a FIFO queue.

\subsection{Proof of eq. (7)}

\begin{proof}
    For a GMM model, the marginal probability density function of $x$ can be written as
    \begin{equation}
        f(x)=\sum_i f(x|z=i)p(x\in C_i)
    \end{equation}
    where $z$ is an indicator that reflects which class $x$ falls in and $C_i$ is the $i^{th}$ class. Since $\sum_i P(x\in C_i) = 1$ by Jensen's inequality we have (note $H(X)=\int(-p(x)\log p(x))dx=\int g(x)dx$ and $g(x)$ is a convex function)
    \begin{equation}
    \begin{split}
    H(X) &= H\left(\sum_i f(x|z=i)p(x\in C_i)\right) \\ & \geq \sum_i p(x\in C_i) H(f(x|z=i))
    \end{split}
    \end{equation}
    In our case we have $p(x\in C_1) = p(x \in C_2)=\frac{1}{2}$, then
    \begin{equation}
    \begin{split}
        H(X) &\geq \frac{1}{2}\left(H(X; \bm{\mu}_1, \bm{\Sigma}_1) + H(X; \bm{\mu}_2, \bm{\Sigma}_2)\right) \\
        & = K_L(X)
    \end{split}
    \end{equation}
    Hence we get a lower bound of $H(X)$ as the right side of the inequality. On the other hand, an upper bound as proposed in \citet{huber2008entropy} is
    \begin{equation}
    \begin{split}
        K_U(X) &= -2\times \frac{1}{2} \times \log\frac{1}{2} + \\
        & \frac{1}{2}\left(H(X; \bm{\mu}_1, \bm{\Sigma}_1) + H(X; \bm{\mu}_2, \bm{\Sigma}_2)\right) \\
        &= 0.693 + K_L(X)
    \end{split}
    \end{equation}
    To summarize
    \begin{equation}
        K_L(X) \leq H(X) \leq K_U(X) = 0.693+K_L(X) 
    \end{equation}
    Then through maximizing the lower bound $K_L(X)$ we can maximize $H(X)$.
\end{proof}

\end{document}